\pdfoutput=1

\documentclass[11pt]{article}

\usepackage[preprint]{acl}

\usepackage{times}
\usepackage{latexsym}
\usepackage{amssymb}
\usepackage[T1]{fontenc}

\usepackage[utf8]{inputenc}

\usepackage{microtype}

\usepackage{inconsolata}

\usepackage{graphicx} %
\usepackage{natbib}  %
\usepackage{caption} %
\usepackage{algorithm}
\usepackage{listings}
\usepackage{algorithmic}
\usepackage{amsmath}
\usepackage{booktabs}
\usepackage{multirow}
\usepackage[outdir=./]{epstopdf}
\usepackage{enumitem}
\usepackage{caption}
\usepackage{subcaption}
\usepackage{subfloat}
\usepackage{newfloat}
\usepackage{graphicx}
\usepackage[normalem]{ulem}
\usepackage{framed}
\usepackage{mdframed}
\usepackage{xcolor}
\usepackage{lipsum}
\usepackage{float}
\usepackage{hyperref}

\definecolor{shadecolor}{gray}{0.9}
\definecolor{LightBlue}{rgb}{0.68, 0.85, 0.90}
\definecolor{LightGreen}{rgb}{0.85, 1, 0.85}

\definecolor{LightYellow}{rgb}{1, 1, 0.71}
\definecolor{LightPurple}{rgb}{0.8, 0.6, 1}

\definecolor{mycolor}{RGB}{68,114,196}
\newcommand{\zy}[1]{\textcolor{black}{#1}}

\newlist{todolist}{itemize}{2}
\setlist[todolist]{label=$\square$}
\usepackage{pifont}

\usepackage{array}
\newcolumntype{L}[1]{>{\raggedright\let\newline\\\arraybackslash\hspace{0pt}}m{#1}}
\newcolumntype{C}[1]{>{\centering\let\newline  \\\arraybackslash\hspace{0pt}}m{#1}}
\newcolumntype{R}[1]{>{\raggedleft\let\newline \\\arraybackslash\hspace{0pt}}m{#1}}

\AtBeginDocument{%
  \providecommand\BibTeX{{%
    \normalfont B\kern-0.5em{\scshape i\kern-0.25em b}\kern-0.8em\TeX}}
}

\author{
 \textbf{Yong Zhao\textsuperscript{*1}},
 \textbf{Kai Xu\textsuperscript{*1}},
 \textbf{Zhengqiu Zhu\textsuperscript{1}},
 \textbf{Yue Hu\textsuperscript{1}},
  \\
 \textbf{Zhiheng Zheng\textsuperscript{2}},
 \textbf{Yingfeng Chen\textsuperscript{2}},
 \textbf{Yatai Ji\textsuperscript{1}},
 \textbf{Chen Gao \textsuperscript{2}},
 \textbf{Yong Li\textsuperscript{2}},
 \textbf{Jincai Huang\textsuperscript{1}}
\\
 \textsuperscript{1}National University of Defense Technology,
 \textsuperscript{2}Tsinghua University, \\
 \textsuperscript{*}Equal contribution
\\
}

\title{
CityEQA: A Hierarchical LLM Agent on Embodied Question Answering Benchmark in City Space
}
\begin{document}
\maketitle
\begin{abstract}
Embodied Question Answering (EQA) has primarily focused on indoor environments, leaving the complexities of urban settings—spanning environment, action, and perception—largely unexplored. To bridge this gap, we introduce \textbf{CityEQA}, a new task where an embodied agent answers open-vocabulary questions through active exploration in dynamic city spaces. To support this task, we present \textbf{CityEQA-EC}, the first benchmark dataset featuring 1,412 human-annotated tasks across six categories, grounded in a realistic 3D urban simulator. Moreover, we propose \textbf{Planner-Manager-Actor (PMA)}, a novel agent tailored for CityEQA. PMA enables long-horizon planning and hierarchical task execution: the \textit{Planner} breaks down the question answering into sub-tasks, the \textit{Manager} maintains an object-centric cognitive map for spatial reasoning during the process control, and the specialized \textit{Actors} handle navigation, exploration, and collection sub-tasks. Experiments demonstrate that PMA achieves 60.7\% of human-level answering accuracy, significantly outperforming competitive baselines. While promising, the performance gap compared to humans highlights the need for enhanced visual reasoning in CityEQA. This work paves the way for future advancements in urban spatial intelligence. Dataset and code are available at \url{https://anonymous.4open.science/r/CityEQA-3027}. 


\end{abstract}



\section{Introduction}
\label{sec::intro}

Embodied Question Answering (EQA) \cite{das2018embodied} represents a challenging task at the intersection of natural language processing, computer vision, and robotics, where an embodied agent (e.g., a UAV) must actively explore its environment to answer questions posed in natural language. \zy{While most existing research has concentrated on indoor EQA tasks \cite{gao2023room, pena2023visual} or traditional indoor/outdoor Visual Question Answering (VQA) tasks \cite{sun20243d}, relatively little attention has been dedicated to EQA tasks in open-ended city space, as shown in Table \ref{table:dataset}.} Nevertheless, extending EQA to city space is crucial for numerous real-world applications, including autonomous systems \cite{kalinowska2023embodied}, urban region profiling \cite{yan2024urbanclip}, and city planning \cite{gao2024embodiedcity}. 

EQA tasks in city space (referred to as CityEQA) introduce a unique set of challenges that fundamentally differ from those encountered in indoor environments. Compared to indoor EQA, CityEQA faces three main challenges: 

1) \textbf{Environmental complexity with ambiguous objects}: 
Urban environments are inherently more complex,  featuring a diverse range of objects and structures, many of which are visually similar and difficult to distinguish without detailed semantic information (e.g., buildings, roads, and vehicles). This complexity makes it challenging to construct task instructions and specify the desired information accurately \cite{ji2025towards,xu2025geonav}. 

2) \textbf{Action complexity in cross-scale space}: 
The vast geographical scale of city space compels agents to adopt larger movement amplitudes to enhance exploration efficiency. However, it might risk overlooking detailed information within the scene. Therefore, agents require cross-scale action adjustment capabilities to effectively balance long-distance path planning with fine-grained movement and angular control.

3) \textbf{Perception complexity with observation dynamics}: 
Observations can vary greatly depending on distance, orientation, and perspective. For example, an object may look completely different up close than it does from afar or from different angles. These differences pose challenges for consistency and can affect the accuracy of answer generation, as embodied agents must adapt to the dynamic and complex nature of urban environments.

\begin{table}
\centering
\caption{\zy{CityEQA-EC vs existing benchmarks.}}
\label{table:dataset}
\renewcommand\arraystretch{1.2}
\resizebox{\linewidth}{!}{
\begin{tabular}{cccccc}
 \hline
              & Platform  & Reference & Place  & Open Vocab & Active  \\ \hline
EQA-v1       & House3D      & \cite{das2018embodied}  & Indoor & \textcolor{red}{\ding{55}}          & \textcolor{green}{\ding{51}}  \\
IQUAD        & AI2-THOR     & \cite{gordon2018iqa} & Indoor & \textcolor{red}{\ding{55}}          & \textcolor{green}{\ding{51}}  \\
MP3D-EQA     & Matterport3D & \cite{wijmans2019embodied} & Indoor & \textcolor{red}{\ding{55}}    & \textcolor{green}{\ding{51}}\\
MT-EQA       & House3D      & \cite{yu2019multi} & Indoor & \textcolor{red}{\ding{55}}          & \textcolor{green}{\ding{51}}  \\
K-EQA        & AI2-THOR     & \cite{tan2023knowledge} & Indoor & \textcolor{red}{\ding{55}}          & \textcolor{green}{\ding{51}}  \\
HM-EQA       & HM3D & \cite{ren2024explore} & Indoor & \textcolor{red}{\ding{55}}          & \textcolor{green}{\ding{51}}   \\
S-EQA        & VirtualHome & \cite{dorbala2024s} & Indoor & \textcolor{red}{\ding{55}}          & \textcolor{green}{\ding{51}}    \\
NoisyEQA         & - & \cite{wu2024noisyeqa} & Indoor & \textcolor{green}{\ding{51}}          & \textcolor{green}{\ding{51}}   \\
OpenEQA         & ScanNet/HM3D & \cite{majumdar2024openeqa} & Indoor & \textcolor{green}{\ding{51}}          & \textcolor{green}{\ding{51}}   \\
 \hline
 
City-3DQA          & - & \cite{sun20243d} & Outdoor & \textcolor{green}{\ding{51}}          & \textcolor{red}{\ding{55}}  \\
EarthVQA        & - & \cite{wang2024earthvqa} & Outdoor & \textcolor{green}{\ding{51}}          & \textcolor{red}{\ding{55}}    \\
Open3DVQA         & - & \cite{zhan2025open3dvqa} & Outdoor & \textcolor{green}{\ding{51}}          & \textcolor{red}{\ding{55}}   \\

CityEQA-EC      & EmbodiedCity & - & Outdoor  & \textcolor{green}{\ding{51}}          & \textcolor{green}{\ding{51}}   \\ \hline
\end{tabular}
}
\end{table}

\begin{figure*}[!htb]
\centering
    \includegraphics[width=0.75\linewidth]{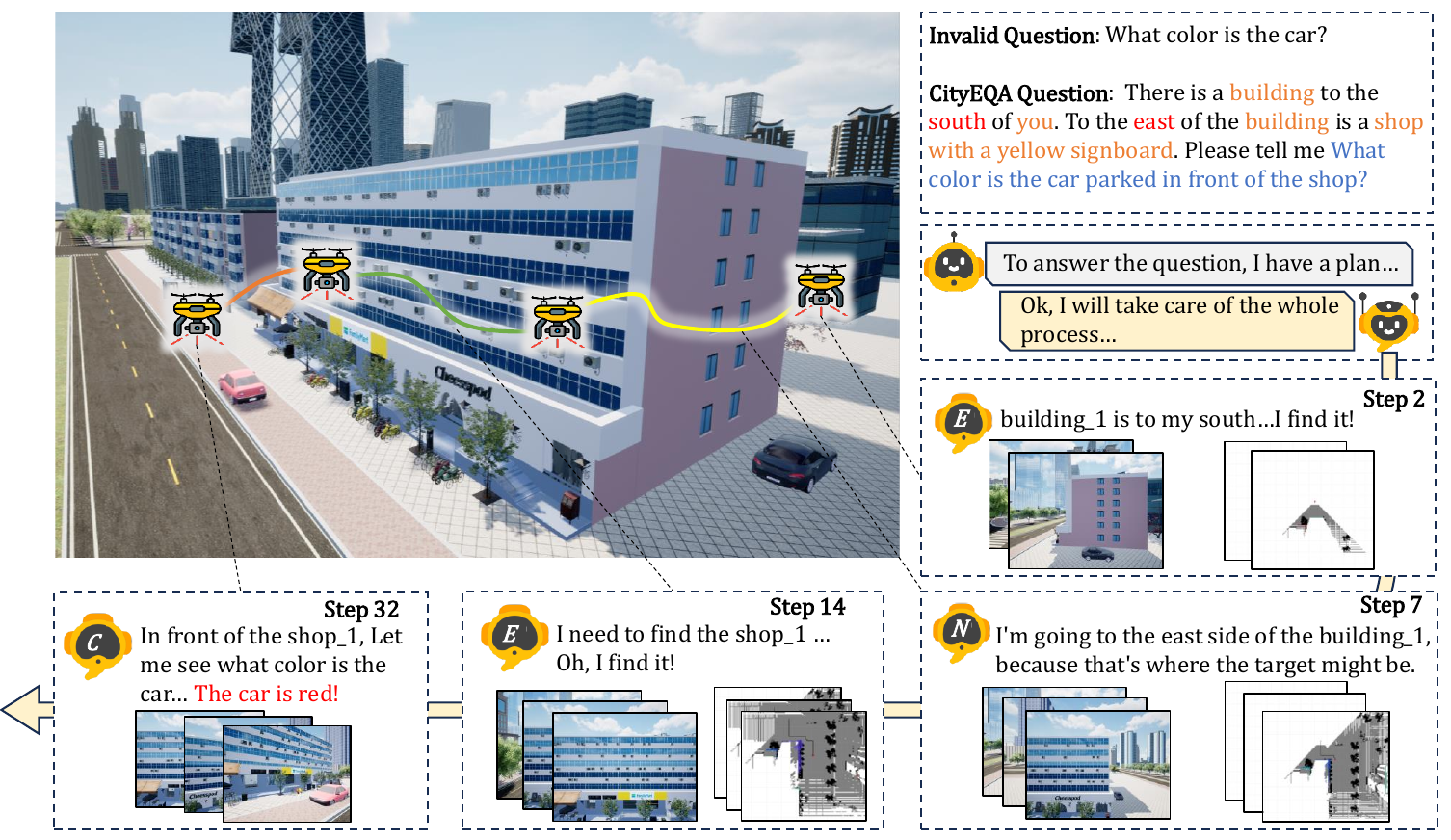}
\caption{The typical workflow of the PMA to address City EQA tasks. There are two cars in this area, thus a valid question must contain landmarks and spatial relationships to specify a car. Given the task, PMA will sequentially complete multiple sub-tasks to find the answer.}
\label{fig:example}
\end{figure*}

As an initial step toward CityEQA, we developed \textbf{CityEQA-EC}, a benchmark dataset to evaluate embodied agents' performance on CityEQA tasks. The distinctions between this dataset and other EQA benchmarks are summarized in Table \ref{table:dataset}. CityEQA-EC comprises six task types characterized by open-vocabulary questions. These tasks utilize urban landmarks and spatial relationships to delineate the expected answer, adhering to human conventions while addressing object ambiguity. This design introduces significant complexity, turning CityEQA into long-horizon tasks that require embodied agents to identify and use landmarks, explore urban environments effectively, and refine observation to generate high-quality answers.

To address CityEQA tasks, we introduce the \textbf{Planner-Manager-Actor (PMA)}, a novel baseline agent powered by large models, designed to emulate human-like rationale for solving long-horizon tasks in urban environments, as illustrated in Figure \ref{fig:example}. PMA employs a hierarchical framework to generate actions and derive answers. The Planner module parses tasks and creates plans consisting of three sub-task types: navigation, exploration, and collection. The Manager oversees the execution of these plans while maintaining a global object-centric cognitive map \cite{deng2024opengraph}. This 2D grid-based representation enables precise object identification (retrieval) and efficient management of long-term landmark information. The Actor generates specific actions based on the Manager's instructions through its components: Navigator, Explorer, and Collector. Notably, the Collector integrates the Vision Language Model (VLM) as its Vision Language Action (VLA) module to refine observations and generate high-quality answers.
PMA's performance is assessed against five types of baselines, including humans. 
Results show that humans perform best in CityEQA, while PMA achieves 60.73\% of human accuracy in answering questions, highlighting both the challenge and validity of the proposed benchmarks. 


\begin{figure*}[!htb]
\centering
    \includegraphics[width=\linewidth]{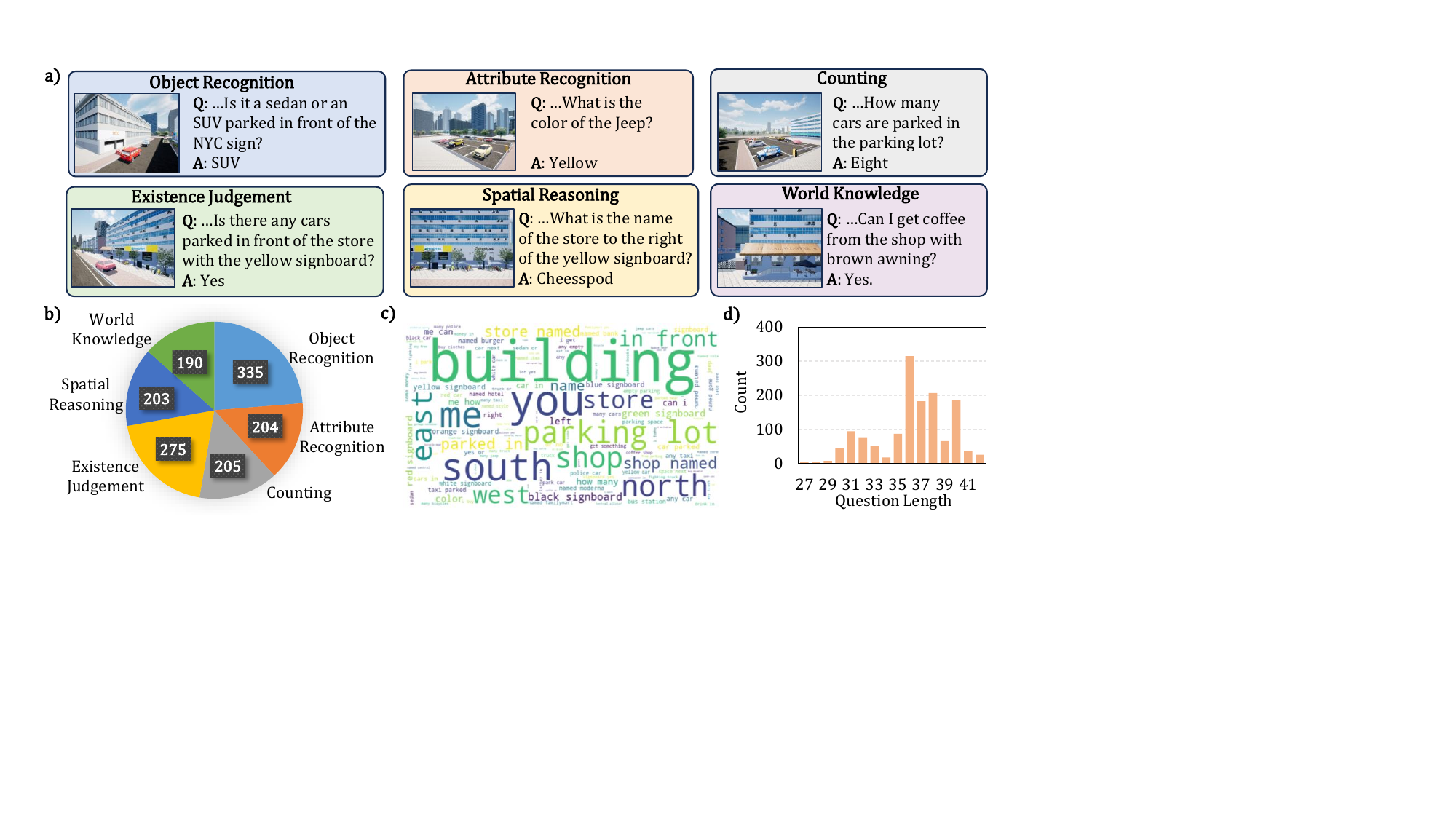}
\caption{\zy{Task examples and dataset statistics of the CityEQA-EC.}}
\label{fig:dataset}
\end{figure*}

In summary, this paper makes the following significant contributions:
\vspace{-8pt}
\begin{itemize}[leftmargin=*]
    \item To the best of our knowledge, we present the first open-ended embodied question answering benchmark for city space, namely CityEQA-EC.
    \vspace{-7pt}
    \item We propose a novel baseline model, PMA, which is capable of solving long-horizon tasks for CityEQA tasks with a human-like rationale.
     \vspace{-7pt}
    \item Experimental results demonstrate that our approach outperforms existing baselines in tackling the CityEQA task. However, the gap with human performance highlights opportunities for future research to improve visual thinking and reasoning in embodied agents for city spaces.
\end{itemize}

\section{CityEQA-EC Dataset}\label{sec::dataset}
\vspace{-0.2cm}
In this section, we outline the formulation of the EQA task and describe the dataset collection process for CityEQA-EC. To address real-world demands, such as urban governance and public services, we draw upon previous research \cite{majumdar2024openeqa, das2018embodied} to define six distinct task types. \zy{Examples and statistics of the dataset are presented in Figure \ref{fig:dataset}. }

\vspace{-0.2cm}

\subsection{Task Formulation}

An instance of the EQA task is defined by the 4-tuple: $\xi =(e,q,y,{{p}_{0}})$, where $e$ is the simulated or real 3D scene that agent can interact with, $q$ is the question, and $y$ is the ground truth answer. The ${{p}_{0}}$ denotes the agent’s initial pose, including 3D position and orientation. Given the instance $\xi $, the goal is for the embodied agent (e.g., drones) to complete the task by gathering the required information from $e$ and generating the answer $\hat{y}$ in response to $q$. Specifically, the agent starts at the initial pose ${{p}_{0}}$ and interacts with the scene $e$ step by step. At each time step $t$, the agent can move to a specific pose ${{p}_{t}}$, and obtain an observation ${{o}_{t}}=(I_{t}^{rgb},I_{t}^{d})$ from the scene, where $I_{t}^{rgb}\in {{\mathbb{R}}^{H\times W\times 3}}$ is the RGB image and $I_{t}^{d}\in {{\mathbb{R}}^{H\times W}}$ is the depth image. Based on these observations, the agent generates the answer $\hat{y}$. The key challenge is to produce a high-quality answer while minimizing the time steps required. 

\subsection{Dataset Collection and Validation}

\zy{To obtain a high-quality dataset, we employed EmbodiedCity} \cite{gao2024embodiedcity}, which is a highly realistic 3D simulation platform based on the buildings, roads, and other elements in a real city. It is implemented using Unreal Engine 4 \cite{sanders2016introduction} and Microsoft AirSim plugins \cite{shah2018airsim}. The collection process is to determine the 4-tuple elements $\xi =(e,{{p}_{0}},q,y)$ of each instance. Unlike indoor simulators with many different scenes, EmbodiedCity is a coherent and extensive scene. As a result, for all instances, their scene $e$ corresponds to EmbodiedCity.

The dataset collection process involves two steps, completed by five human annotators. The first step is raw Q\&A generation, where raw questions and answers are created. The second step is task supplementation, which includes determining the agent's initial pose and and refining the question descriptions accordingly. Once these steps are completed, the dataset undergoes validation and filtering. More details can be found in Appendix \ref{a_data_collection}.  

\paragraph{Raw Q\&A Generation} 

We instructed human annotators to explore the EmbodiedCity environment freely and generate question-answer pairs based on their observations of RGB images. The raw questions ${q^r}$ and answers $y$ are presented as open-vocabulary text. In addition to documenting the question-answer pairs, annotators were also required to record the pose ${p^{obs}}$ from which the RGB images were captured, along with the pose ${p^{tar}}$ of the target object referenced in each question. These information can be leveraged for a comprehensive evaluation of the agent's performance. After basic revision process, we have finally collected a total of 443 such instances, with each raw task instance denoted as ${\xi ^r} = ({q^r}, y, {p^{obs}}, {p^{tar}})$.

\paragraph{Task Supplementation} 

Building upon the raw task instances, we further established the agent's initial pose and refined the questions accordingly. For each raw task, the initial pose \( p_{0} \) of the agent was set within a 200-meter range of the target object's pose \( p^{tar} \). Given the complexity of urban environments, and to ensure that each expected answer is unique, we enriched the questions with descriptions based on landmarks. An example of this process is illustrated in Figure \ref{fig:example}. For each raw task, we generated at least four distinct initial poses and transformed each raw question into at least four different inquiries. Ultimately, this process yielded a total of 2,212 task instances.

\paragraph{Dataset Validation} 

Each task instance created by human annotators was rigorously evaluated by two independent human reviewers. These reviewers were responsible for determining whether the questions posed were answerable and clear, as well as verifying the uniqueness and accuracy of the target objects and their corresponding answers. Any task instance identified with issues was excluded. The final dataset comprises 1,412 task instances, with detailed statistics presented in Figure \ref{fig:dataset}.




\section{PMA: A Hierarchical LLM Agent for CityEQA Task}\label{method}

 \begin{figure*}[!htb]
\centering
    \includegraphics[width=0.8\linewidth]{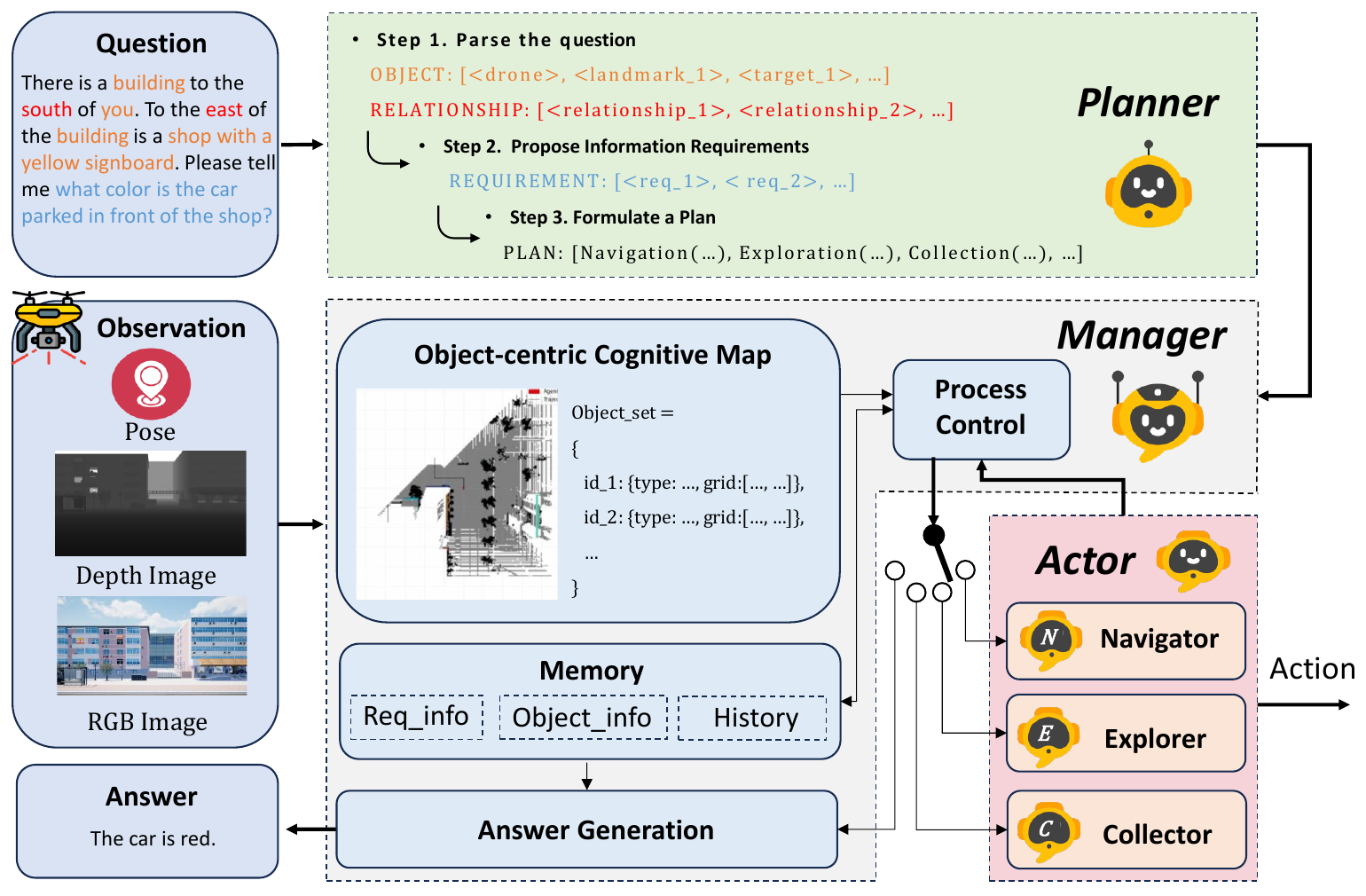}
\caption{The overview of our proposed PMA agent.}
\label{fig:framework}
\end{figure*}

\subsection{Overview}

\zy{An overview of the proposed PMA agent for CityEQA tasks is shown in Figure \ref{fig:framework}. The PMA comprises three major modules: Planner, Manager, and Actor, all powered by pre-trained foundation models. Planner is responsible for parsing the question $q$ and formulating an executable plan before any actions are taken. Manager serves as the core module, receiving structured information from Planner and processing observations at each time step to maintain an object-centric cognitive map using an VLM. Additionally, through a process control module, Manager issues task instructions to Actor, which then utilizes various action generators to execute the required responses. Once the plan is completed, Manager generates an answer based on  its accumulated memory.}

\subsection{Planner Module}
The question descriptions in CityEQA tasks contain extensive information, including several objects, spatial relationships, and the information that needs to be collected. To address the open-ended question descriptions, we leveraged pre-trained LLMs and designed a few-shot prompt that employs a three-step Chain of Thought (CoT) reasoning \cite{wei2022chain} to parse the question and formulate a plan.

As illustrated in Figure \ref{fig:framework}, all objects and spatial relationships mentioned in the question are first extracted. Simultaneously, the information necessary to answer the question is identified as corresponding requirements. Based on these requirements, a plan is created consisting of three distinct types of sub-tasks: (1) \textbf{Collection sub-tasks} gather the requisite information, (2) \textbf{Exploration sub-tasks}  identify landmarks or target objects, and (3) \textbf{Navigation sub-tasks} enable efficient access to specific areas, thereby narrowing the exploration scope. To ensure the plan is executable, we have developed several strategies to guide the LLMs, with details provided in Appendix \ref{a_planner}.

\subsection{Manager Module}

The Manager possesses the capability to oversee and manage the gradual implementation of long-term plans. This is made possible by its Memory module and Map module, which facilitate the organized storage of observations and track execution progress as the plan unfolds.

\paragraph{Object-Centric Cognitive Map} 
The object-centric cognitive map takes the initial pose of the agent as the origin, uses 2D grids to discretize the surrounding environment, and  records the distribution of landmark objects based on grid indices. The map at time step $t$-1 is represented as ${M_{t{\rm{ - }}1}}{\rm{ = }}\{ obj\_1,{\rm{ }}obj\_2,{\rm{ }}...\} $, where the $obj\_1$ and $obj\_2$ are the object IDs corresponding to specific objects in the environment. At each time step $t$,  the agent leverages egocentric observations represented as ${{o}_{t}}=(I_{t}^{rgb},I_{t}^{d})$ to construct the added map ${{m}_{t}}$ to record the landmark objects appeared at current observation, denoting as ${{m}_{t}}=Construct({{o}_{t}, p_t})$. To implement the functionality of $Construct()$, we utilized the GroundSAM model \cite{bousselham2024grounding} for grounding and segmenting landmark objects from $I_{t}^{rgb}$. By integrating pose information with depth data from $I_{t}^{d}$, we can obtain a 3D point cloud representation of these objects, subsequently projected onto 2D grids. After denoising and filtering, we obtained the finalized added map, denoted by ${{m}_{t}}$.

The added map ${{m}_{t}}$ will be fused with the ${{M}_{t\text{-}1}}$ by merging the same object observed at different time steps, so objects are guaranteed to be unique in the map, denoting as ${{M}_{t}}=Merge({{m}_{t}},{{M}_{t-1}})$. More details can be found in Appendix \ref{occm}.

\paragraph{Other Modules}
Memory module records important information in the perceptual process, which mainly includes three aspects. \textit{Req\_info} records the collected information, and \textit{Object\_info} records object information, such as the object's ID in the map. \textit{History} records the completion progress of sub-tasks and the execution results of actions.

Process Control is designed to determine the next sub-task to be executed based on the current progress of the plan. It also serves as the interface for interaction with the Actor. Once all sub-tasks in the plan have been completed, Process Control invokes the Answer Generation module to produce the final response. The Answer Generation process is also driven by LLMs, employing a zero-shot prompt specifically crafted to generate answers based on the \textit{Req\_info} stored in memory.

\subsection{Actor Module}
To address the distinct objectives of the three types of sub-tasks, we introduce three specialized low-level action generators: Navigator, Explorer, and Collector. The Navigator and Explorer rely on distinct deterministic policies to generate actions based on the cognitive map. In contrast, the Collector uses a VLA policy, which directly derives actions from RGB images. These action models serve as fundamental baselines and provide a foundation for future research enhancements.

\paragraph{Navigator}
The navigation sub-task instructions specify a landmark and a directional relationship. For instance, \textit{Navigation(building\_1, west)} indicates that building\_1 serves as the landmark, with navigation directed to the west of it, where the target object is likely located. The Navigator identifies the nearest navigation point on the map by analyzing the landmark's distribution in conjunction with its spatial relationship. It then employs the A* algorithm to plan a path from the agent's current position to this navigation point. Given the potential incompleteness of recorded landmarks on the map, a multi-step approach is adopted, restricting each step's path length ${L^{nav}}$ to 10 meters. The navigation point is updated following each cognitive map update.

\paragraph{Explorer}
The typical exploration sub-task is described as \textit{Exploration(building\_1, west, red\_car)}, which means the goal is to explore the west side of \textit{building\_1} to find a red car. The explorer uses the Move and Look Around strategy due to the complexity of outdoor environments, where re-observing previously explored areas from different angles can yield different results. The exploration area is defined on the map based on landmark distribution and spatial relationships. A set of exploration points is generated within this area, maintaining a fixed distance of ${L^{exp}=10}$ meters between them. At each point, the agent thoroughly observes its surroundings by looking in four directions: front, back, left, and right. After completing observations at one point, the agent moves to the next closest point and continues until either the target object is found or all points are covered. A VLM is employed to determine whether the target appears in any given observation.

\paragraph{Collector}
The collection sub-task instructions only include an information requirement. We provide a VLM-driven Collector to gather the required information from observations. Additionally, the Collector can select an action from a predefined action set to fine-tune its observation view, enabling the collection of higher-quality information. More details of Collector is presented in Appendix \ref{a_collector}.
\section{Experiment}
\label{expriment}





\subsection{Experiment setup}

\paragraph{Evaluation Metrics}
In CityEQA, we adopted three widely used metrics for evaluating EQA tasks \cite{das2018embodied}: Question Answering Accuracy (QAA) assesses the correctness of the answers by comparing them to the ground truth. The open-vocabulary nature of the CityEQA task poses challenges for evaluation. Inspired by OpenEQA \cite{majumdar2024openeqa}, we employed an LLM as the judge to assign scores $\theta  \in \left\{ {1,2,...,5} \right\}$ to the answers. For detailed information, please refer to the Appendix \ref{metric}. Navigation Error (NE) is measured by the distance between the agent's final position and the target object ${p^{tar}}$ upon task completion, reflecting whether the agent successfully located and approached the target. Mean Time Step (MTS) is calculated as the average number of time steps required to complete all tasks, indicating the efficiency of the embodied agent's action strategy.

\paragraph{Implementation Details}

\zy{For each task, the object-centric cognitive map is constructed centered around the agent's initial pose, with a side length of 400 meters and a resolution of 1 meter. The dimension of the images obtained by the agent is 640×480, and we considered buildings as landmarks and accounted for four spatial relationships: north, south, east, and west. Additionally, the total number of time steps for navigation and exploration is limited to 50 steps and the maximum steps for collection is 10. GPT-4o and GPT-4 are the default  VLM and LLM used in the PMA. Due to API limitations, 200 tasks are randomly selected from CityEQA-EC for the experiments.}

\paragraph{Baselines}

\zy{We compare various models in a zero-shot setting, including five categories of baselines that are widely used in studies of EQA tasks. More details of baselines can be found in Appendix \ref{a_baseline}.}
\begin{itemize}[leftmargin=*]
    \item \textbf{Blind Agents \cite{majumdar2024openeqa}} \zy{generate answers based solely on the text of questions without obtaining any visual inputs. It serves as a reference for assessing the extent to which one can rely purely on prior world knowledge and/or random guessing.}
    \vspace{-6pt}
    
    \item \textbf{Socratic Agents \cite{jiang2025beyond}} \zy{use the VLM (GPT-4o) to convert the visual input during the exploration process into image captions, and then uses LLMs to generate answers based on these descriptions.}
    \vspace{-6pt}

    \item \textbf{VQA Agents} \zy{bypass the active exploration process and is directly provided with the RGB image obtained from the ${p^{obs}}$ to answer the questions. This approach aims to assess the visual perception and reasoning capabilities of VLMs in urban environments, while eliminating the interference of embodied actions.}
    \vspace{-6pt}
    
    \item  \textbf{Exploring Agents \cite{ren2024explore}} \zy{actively acquire visual inputs using Random Exploration (RE) and Frontier-Based Exploration (FBE), both commonly used as indoor baselines.}
    
    \vspace{-6pt}
    \item \textbf{Human Agents} \zy{are employed to establish human-level performance metrics on our benchmark. We categorize human agents as H-VQA or H-EQA, depending on whether they actively acquire visual inputs.}
    

\end{itemize}

\subsection{\zy{Comparison with State-of-the-art}}


\begin{table}
\centering
\caption{\zy{Performance of baselines and the proposed PMA on the CityEQA tasks.}}
\label{table:comparison}
\renewcommand\arraystretch{1.2}
\resizebox{\linewidth}{!}{
\begin{tabular}{lccc}
\hline
\multicolumn{1}{c}{} & QAA (1-5) $\uparrow$     & NE (m) $\downarrow$         & MTS $\downarrow$        \\ \hline
\colorbox{LightBlue}{Blind Agents}          &           &             &             \\
GPT-4             & 1.90±1.64 & -           & -           \\
Qwen-2.5          & 2.34±1.88 & -           & -           \\ 
LLaMA-v3.1-8b     & 2.31±1.72 & -           & -           \\
DeepSeek-v3       & 2.03±1.41 & -           & -           \\
\hline
\colorbox{LightGreen}{Socratic Agents (VLM/LLM)}           &           &             &             \\
GPT-4o/GPT-4          & 2.71±1.72 & -           & -           \\
GPT-4o/Qwen-2.5       & 2.77±1.49 & -           & -           \\ 
GPT-4o/LLaMA-v3.1-8b  & 2.70±1.71 & -           & -           \\
GPT-4o/DeepSeek-v3    & 2.82±1.53 & -           & -           \\ 
\hline
\colorbox{pink}{VQA Agents}           &           &             &             \\
GPT-4o            & 4.37±1.35 & -           & -           \\
Qwen-2.5          & 4.00±1.67 & -           & -           \\ 
LLaVA-v1.5-7b     & 3.81±2.01 & -           & -           \\
\hline
\colorbox{LightYellow}{Exploring Agents}           &           &             &             \\  
RE      & 2.19±2.64 & 73.31±45.43 & 46.41±10.41 \\ 
FEB     & 2.31±2.54 & 86.92±53.71 & 39.31±32.17 \\
\hline
\colorbox{LightPurple}{Human Agents}               &           &             &             \\
H-VQA             & 4.87±0.72 & -           & -           \\
H-EQA             & 4.94±0.21 & 38.72±40.87 & 9.31±6.32   \\ \hline
\colorbox{shadecolor}{PMA (ours)}                & 3.00±1.96 & 46.56±36.39 & 24.44±14.39 \\ \hline
\end{tabular}
}
\end{table}

\zy{As shown in Table \ref{table:comparison}, human agents in both VQA and EQA settings achieve the highest QAA scores—4.87±0.72 for H-VQA and 4.94±0.21 for H-EQA—representing the upper bound for answer quality. They also demonstrate exceptional efficiency, with the lowest navigation error (38.72±40.17m) and completion steps (9.31±6.32), setting the gold standard for both quality and efficiency.}

\zy{For automated methods, VQA agents like GPT-4o reach QAA scores up to 4.37±1.35, approaching human performance in answer quality, but lack active exploration abilities, preventing assessment of their overall task efficiency. Blind and Socratic agents perform significantly worse, with QAA between 1.90 and 2.82, showing the shortcomings of methods without visual information or with only language-based reasoning.}

\zy{Exploring agents such as RE and FEB can handle active exploration and answering, but their QAA scores are low (2.19–2.31) and their NA and MTS are much higher (e.g., FEB: 86.92±53.71m, 39.31±32.17), resulting in less effective execution. In contrast, PMA achieves a balanced performance: its QAA of 3.00±1.96 is higher than all exploring, blind, and Socratic agents, though still just 60.73\% that of H-EQA. Importantly, PMA’s navigation error (46.56±36.39m) and completion step (24.44±14.39) are dramatically less than traditional exploring agents, demonstrating notable practical gains.}

Overall, the comparison with baselines reveals that \textit{accurate visual inputs and reasoning are crucial for improving performance in CityEQA tasks}. Additionally, obtaining accurate visual inputs \textit{relies on the efficient exploration using landmarks and spatial relationships} in urban environments. 

\subsection{\zy{Ablation Studies}}
\zy{We conduct ablation studies on the Object-Centric Cognitive Map, navigator, and explorer modules in PMA, as shown in Table \ref{table:ablation}. Removing any of these modules leads to a significant decline in performance. Without the map, the agent becomes confused by similar landmarks in the environment and fails to perform effective active perception, resulting in the worst ablation outcome. Furthermore, the absence of the navigator is more detrimental than that of the explorer, further highlighting the importance of landmark-based navigation in urban environments.}

\begin{table}
\centering
\caption{\zy{Ablation results.}}
\label{table:ablation}
\renewcommand\arraystretch{1.2}
\resizebox{\linewidth}{!}{
\begin{tabular}{lccc}
\hline
\multicolumn{1}{c}{} & QAA (1-5) $\uparrow$     & NE (m) $\downarrow$       & MTS $\downarrow$        \\ \hline
PMA w/o map        & 2.31±1.82 & 76.41±48.64  & 43.27±31.92           \\
PMA w/o navigator  & 2.33±1.64 & 68.31±46.91 & 38.83±27.71           \\ 
PMA w/o explorer   & 2.68±1.87 & 57.13±41.43  & 20.62±15.11           \\
PMA       & 3.00±1.96 & 46.56±36.39 & 24.44±14.39    \\
\hline
\end{tabular}
}
\end{table}

\subsection{Effectiveness of Collector Module}

\begin{figure}[!htb]
\centering
    \includegraphics[width=0.9\linewidth]{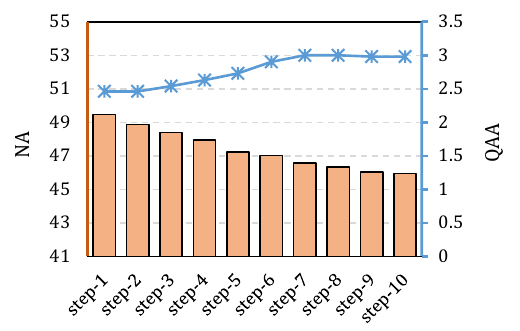}
\caption{The performance of the Collector module at different steps.}
\label{fig:pma}
\end{figure}

This section further investigates the effectiveness of the collector module, specifically the impact of fine-grained observation adjustments on performance. We recorded the observation at each step (10 steps in total) during the collection phase and calculated relevant metrics, as shown in Figure \ref{fig:pma}. 

\zy{\textit{The Collector significantly affects outcomes}: as steps increase, NE decreases and QAA rises, helping the agent approach targets and improve accuracy. However, QAA plateaus, with Step 10 slightly lower than Step 9, possibly due to "over-adjustment" degrading visual input quality.}

We further analyzed the Collector's taken actions, as detailed in Appendix \ref{collector_action}. The most frequent action was \textit{KeepStill}, reflecting effective Navigation and Exploration sub-tasks that help the agent successfully approach the target object. Additionally, the proportions of \textit{MoveForward}, \textit{TurnLeft}, and \textit{TurnRight} were also relatively high. 
Case analysis revealed that when a target object enters the agent’s view, it tends to stop, possibly cause the object too far away or only partially visible. In such instances, the agent must either \textit{MoveForward} to reduce distance or use \textit{TurnLeft} and \textit{TurnRight} to adjust its orientation for better observation and information gathering about the target object. However, these adjustments remain limited, as illustrated in two cases presented in Appendix \ref{collector_action}.

\section{Related Works}
\label{related}

\subsection{QA and EQA}
\zy{Early research on using language to guide perception from given input is known as Question Answering (QA), such as Visual QA (VQA) \cite{ishmam2024image} and 3DQA \cite{zhan2025open3dvqa}. These QA tasks require agents to answer questions based solely on provided information (images or cloud points) \cite{chandrasegaranhourvideo}. In contrast, EQA involves agents actively exploring within an environment to seek visual inputs and enhance answer reliability \cite{das2018embodied}. Due to cost and hardware limitations, several virtual indoor simulators have been developed for EQA tasks \cite{liu2024aligning}, resulting in indoor-focused datasets such as EQA-v1 \cite{das2018embodied} and MT-EQA \cite{yu2019multi}. However, although there are already several QA task datasets for outdoor environments, such as City-3DQA \cite{sun20243d} and Open3DVQA \cite{zhan2025open3dvqa}, EQA tasks have yet to be extended to outdoor settings, as shown in Table \ref{table:dataset}.}

Recently, urban environment simulators like EmbodiedCity \cite{gao2024embodiedcity}, CityNav \cite{lee2024citynav}, and AerialVLN \cite{liu2023aerialvln} have emerged, though they mainly focus on navigation task. \zy{EmbodiedCity provides an urban EQA dataset, but it functions more like VQA and ignores the active perception.} 
Moreover, due to the limited generalization capabilities of models at the time, only simple questions about basic attributes of objects were considered in these indoor datasets \cite{ren2024explore}. However, with the continuous improvement in the understanding and reasoning capabilities of pre-trained VLMs for visual inputs, several open-ended EQA datasets have recently been released, such as Express-bench \cite{jiang2025beyond} and OpenEQA \cite{majumdar2024openeqa}. In comparison, this paper is the first to study the EQA tasks in city space and introduces the benchmark CityEQA-EC --- a high-quality dataset featuring diverse, open-vocabulary questions.

\subsection{LLMs-driven Embodied Agents}

The indoor EQA tasks mainly involve exploration and answer generation sub-tasks \cite{ren2024explore}. In early work \cite{duan2022survey, das2018embodied, lu2019vilbert}, the two sub-tasks are mainly addressed by building and fine-tuning various deep neural networks. Recently, researchers attempt to utilize pre-trained LLMs to solve EQA tasks without any additional fine-tuning \cite{mu2024embodiedgpt, xiang2024language, huang2024manipvqa}. \zy{NaviLLM employed a scheme-based instruction that flexibly casts various tasks into generation problems, including the EQA task \cite{zheng2024towards}.} OpenEQA employed a Frontier-Based Exploration (FBE) strategy for indoor environment exploration and tested the performance of various VLMs on the answer generation \cite{majumdar2024openeqa}. Besides, VLMs was also used to determine which room to explore in indoor environment based their commonsense reasoning capabilities \cite{yinsgsg-nav}. 

These agents, however, cannot be directly used for CityEQA tasks. Unlike indoor spaces, which are confined and divided into rooms, city spaces are vast and open. Agents in cities must navigate using landmarks and spatial relationships for long-term exploration \cite{liu2024navagent}. The proposed PMA addresses this by breaking down and planning for long-horizon CityEQA tasks, using large models across multiple modules to effectively handle open-ended questions and unseen environments.

\section{Conclusion}\label{sec::conclusion}

This paper pioneers the exploration of EQA tasks in outdoor urban environments. First, we introduced CityEQA-EC, the inaugural open-ended benchmark for CityEQA, comprising 1,412 tasks divided into six distinct categories. Second, we proposed a novel agent model (the PMA), designed to tackle long-horizon tasks through hierarchical planning, sensing, and execution.  Experimental results validated the effectiveness of PMA, achieving 60.73\% accuracy relative to human performance and outperforming traditional methods such as the FBE Agent. Nevertheless, challenges remain, including efficiency discrepancies (24.44 vs. 9.31 mean time steps taken by humans) and limitations in visual thinking capabilities. Future research could focus on enhancing PMA with self-reflection and error-correction mechanisms to mitigate error accumulation that can arise in long-horizon tasks.

\clearpage

\section{Limitations}
The work primarily focuses on object-centric question-answering tasks, such as identifying specific objects (e.g., buildings, vehicles) within city spaces.  Further, while our approach is effective for tasks involving static physical entities, it overlooks the importance of social interactions and dynamic events, which are also critical in urban settings.  For instance, questions related to dynamic events (e.g., "Is there a traffic jam on Main Street?"), or environmental conditions (e.g., "Is the park crowded right now?") are not considered up to now. These types of questions require some different sets of reasoning capabilities, such as temporal reasoning, event detection, and social context understanding, which are not currently supported by the Planner-Manager-Actor (PMA) agent.  Future work should expand the scope of CityEQA to include these non-entity-based tasks, further extending PMA and enabling embodied agents to handle a broader range of urban spatial intelligence challenges.




\section{Ethics Statement}
In the data collection, we ensure there is no identifiable information about individuals (faces, license plates) or private properties.
Thus, there is no ethical concern.


\bibliographystyle{ACM-Reference-Format}
\bibliography{bibliography}

\clearpage

\appendix
\section{Appendix}

\subsection{Dataset Collection and Validation}
\label{a_data_collection}
\begin{figure}[!htb]
\centering
    \includegraphics[width=\linewidth]{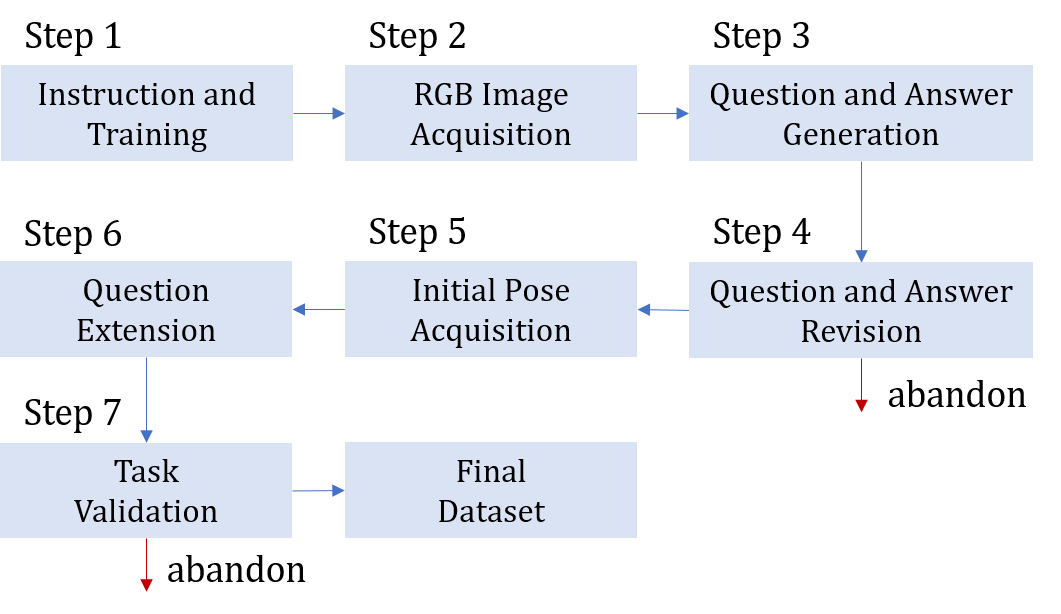}
\caption{The collection and validation process of the CityEQA dataset.}
\label{fig:dataset_collection}
\end{figure}

The collection and validation process of the CityEQA-EC dataset is shown in Figure \ref{fig:dataset_collection}, including Initialization (Step 1), Raw Q\&A Generation (Step 2 to 4), Task Supplementation (Step 5 to 6), and Dataset Validation (Step 7).

In the initialization phase, human annotators were provided with comprehensive briefings and training, during which they were introduced to six distinct types of tasks. Subsequently, in the raw question-and-answer generation stage, annotators were randomly placed within the environment, allowing them to move freely and explore in order to generate questions and answers. Additionally, both the target pose ${p^{tar}}$ and observed pose ${p^{obs}}$ were recorded manually. Then, each question-answer pair was then reviewed by two additional annotators to identify specific issues: (1) Task Duplication, indicating that a similar instance had already been collected; (2) Task Invalidity, meaning that there was no match between the question and answer based on the image. Any tasks identified as problematic were discarded. Furthermore, to ensure the accuracy of pose annotations, we randomly selected 20\% of raw task examples for two rounds of verification regarding their pose annotations.

In the task supplementation phase human annotators were asked to add the initial pose for the task and expand the question.  Buildings are primarily used as landmark objects to expand the question. Then, in the validation stage, each task was independently evaluated by two human reviewers. The details of the review policy are as follows:

\begin{itemize}
    \item Spelling and grammar check is conducted.
    \item The target object must be uniquely identifiable based on descriptions of landmarks and spatial positions.
    
    \item The distance between the initial pose and the target pose must be less than 200 meters.
    
    \item The initial pose is located at a movable position rather than within an obstacle.

\end{itemize}

\zy{Any tasks identified as problematic were removed. To ensure the annotation consistency in the data collection and validation, we conducted Kappa statistical analyses for the raw annotation data from both Question and Answer revision phase and the task validation phase. The Kappa coefficients $\kappa$ for the two phases were 0.93 and 0.89, respectively, indicating a high level of agreement among annotators.}

\subsection{PMA Agent Details}

\paragraph{Details of Planner}
\label{a_planner}
We present the detailed CoT used by the Planner here.

\begin{figure*}[!htb]
\centering
    \includegraphics[width=\linewidth]{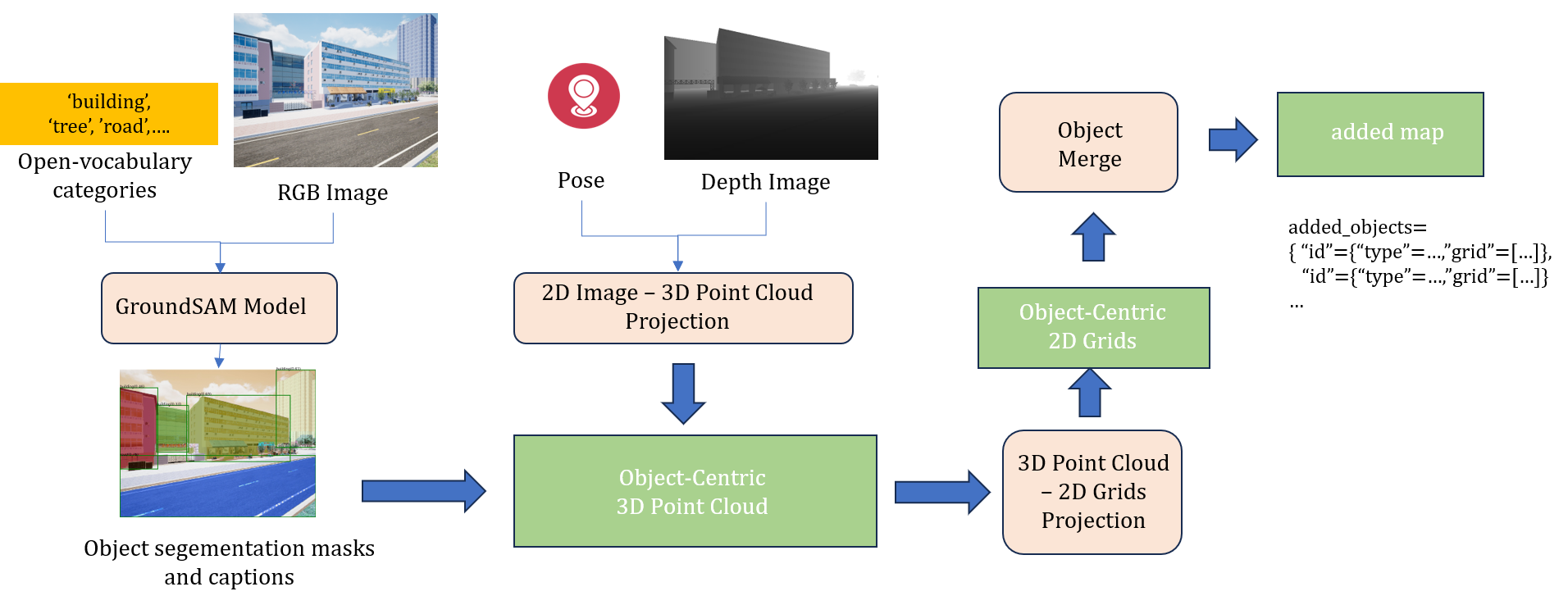}
\caption{The workflow of the construction of the added map.}
\label{fig:cosntruction}
\end{figure*}

\textbf{Step 1}. All the objects mentioned in the question are extracted, along with the spatial relationships between them. Each object is assigned a unique identifier to ensure distinction. Additionally, the state of each object is marked as \texttt{Unknown} as their locations remain uncertain. The agent itself is treated as a special object, with its state marked as \texttt{Known}, allowing it to serve as a unique initial landmark.

\textbf{Step 2}. The information necessary to answer the question is extracted as corresponding information requirements. This step forms the purpose for the following plan generation, as the entire perception process is driven by the need to gather this critical information.

\textbf{Step 3}. An executable plan is formulated by combining three types predefined sub-tasks based on information requirements. To guide LLMs reasoning and constructing an executable plan, we establish a set of simple rules. First, collecting information requires the Collection sub-task. However, before executing this sub-task, the states of the relevant objects must be \texttt{Known}, meaning the objects must already have been located in the environment. Second, the Exploration sub-task can transition an object's state from \texttt{Unknown} to \texttt{Known}. Third, before performing Exploration, the Navigation sub-task can be employed to leverage a \texttt{Known} object as the landmark, enabling the agent to efficiently reach specific locations. This sub-task can reduce the exploration scope and enhances overall efficiency.

\paragraph{Details of Object-Centric Cognitive Map}
\label{occm}

The processing procedure of the function $ Construct()$ is illustrated in Figure \ref{fig:cosntruction}. Firstly, the GroundSAM model is utilized to process the RGB image to obtain object segmentation masks and captions. Meanwhile, the pose and depth image are combined with the camera intrinsic parameters to obtain 3D point cloud data. Then, these two data are merged to obtain the object-centric 3D point cloud. Further, this data is projected onto a 2D grid, and the point cloud data outside the map range is filtered out to obtain the object-centric 2D grids. Finally, objects with repetitive grids are fused to obtain the object-centric added map.

The purpose of the function $ Merge()$ is to fuse the added objects in added map into the global map. This is to ensure that the same object observed from different views is uniquely recorded and retrieved on the map. Therefore, for each added object, we first determine whether the distribution of the object overlaps or is adjacent to any object in the global map. If so, the two objects are merged; if not, the object is directly added to the global map. This paper adopts a simple and effective strategy to determine whether objects are adjacent: when at least one pair of grids in which the two objects are distributed are adjacent, they are considered to have an adjacent relationship. Additionally, it should be noted that multiple object merges may occur in the same round, so the merged object needs to be judged against all other objects in the global map in another round.

\begin{figure}[!htb]
\centering
    \includegraphics[width=\linewidth]{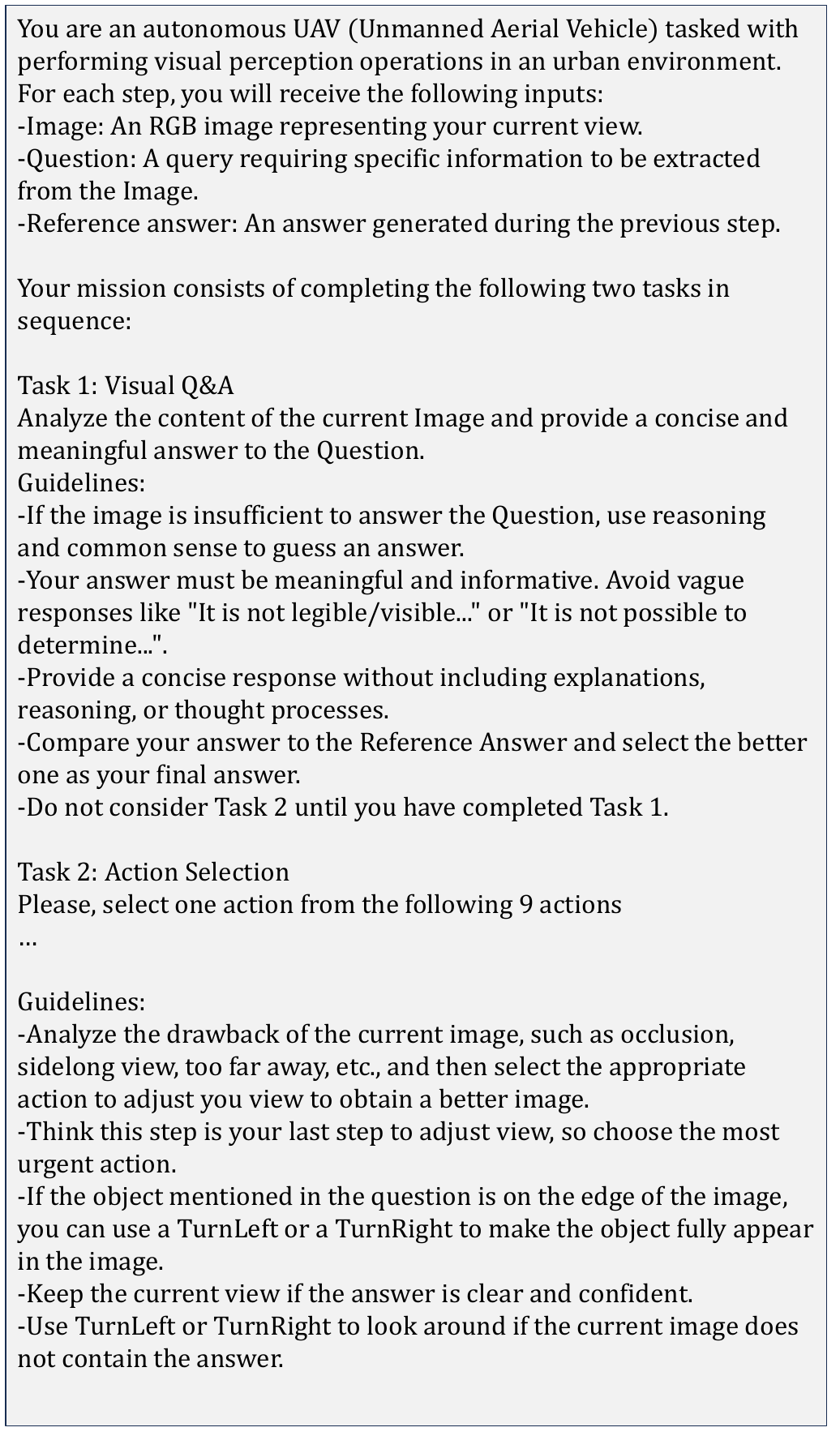}
\caption{The prompt used for Collector.}
\label{fig:prompt_collector}
\end{figure}

\paragraph{Details of Collector}
\label{a_collector}
The prompt provided for MM-LLM in Collector is presented in Figure \ref{fig:prompt_collector}. The Collector needs to complete two tasks in sequence. The first is the VQA task, which involves answering the corresponding questions based on the provided RGB image. The second is action selection, which requires choosing an appropriate action from a discrete set of actions to adjust the observation. The action set used in this study includes \{\textit{MoveForward}, \textit{MoveBack}, \textit{MoveLeft}, \textit{MoveRight}, \textit{MoveUp}, \textit{MoveDown}, \textit{TurnLeft}, \textit{TurnRight}, \textit{KeepStill}\}.

\subsection{Experiments Details}

\paragraph{LLM Scoring}
\label{metric}

\begin{figure}[!htb]
\centering
    \includegraphics[width=\linewidth]{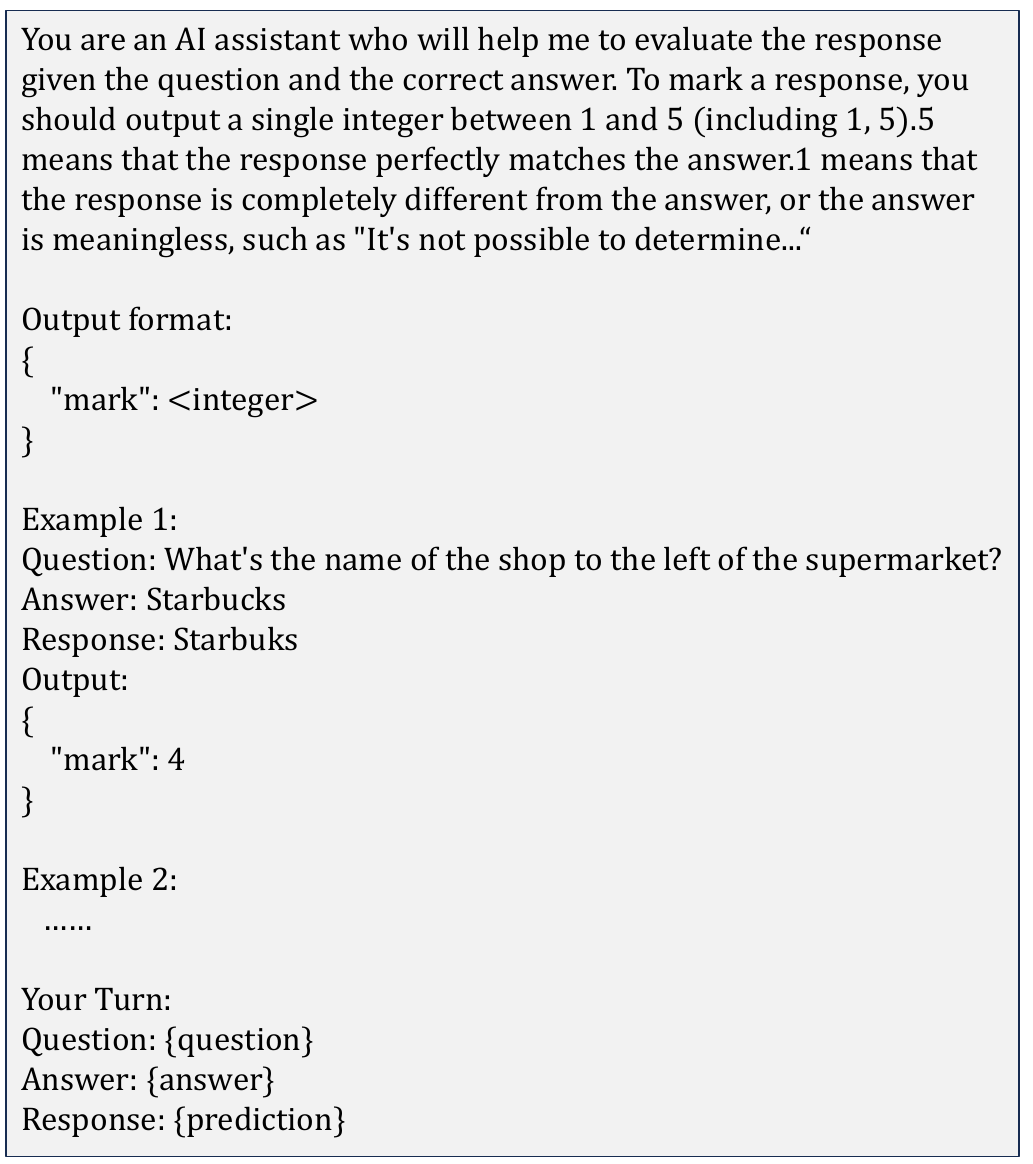}
\caption{The prompt used for LLM scoring.}
\label{fig:llmscore}
\end{figure}

For QAA, we designed an LLM-based automated scoring method by referring to the LLM-Match mechanism in OpenEQA \cite{majumdar2024openeqa}. We show the designed prompt for LLM in Figure \ref{fig:llmscore}.

To investigate the validation of using the LLM as judge, a double blind study is conducted. We randomly sampled 100 answers from the results including the answer generated by the 4 baselines and PMA. Then 2 human evaluators are required to provide their score of the answers while using the prompt in Figure \ref{fig:llmscore} as the task instruction. Since the distribution of scores did not conform to a normal distribution, Spearman's correlation analysis was adopted. The results indicated a significant positive correlation between the scores given by human evaluators and those by LLM judges ($R_s$ = 0.85, $p$ = 0.002). This suggests that using LLMs as judges can effectively evaluate open-ended question-answering results and align well with human judgments.

\begin{figure}[!htb]
\centering
    \includegraphics[width=\linewidth]{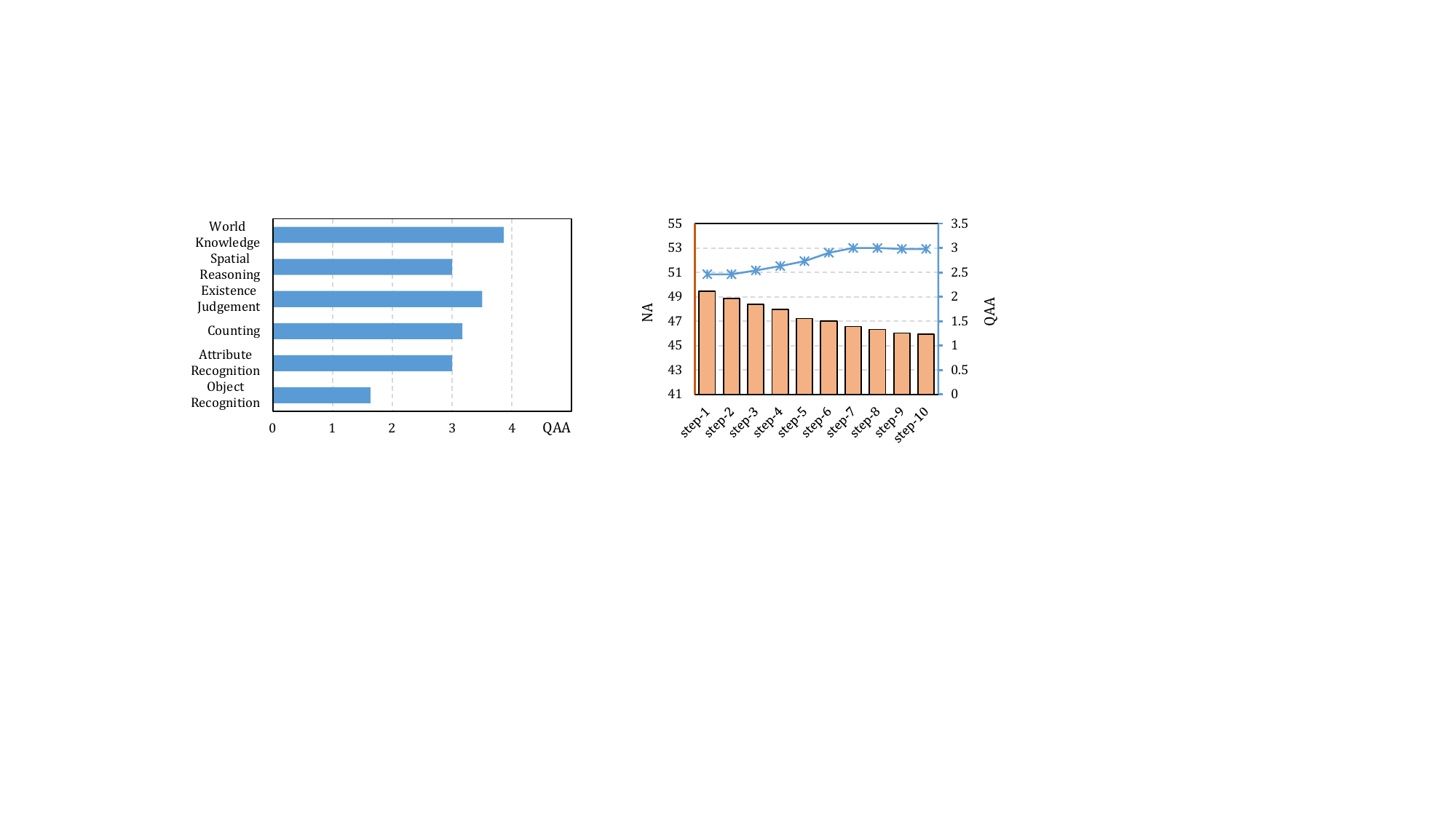}
\caption{Categroy-level performance of the proposed PMA.}
\label{fig:category}
\end{figure}

\begin{figure*}[!htb]
\centering
    \includegraphics[width=0.9\textwidth]{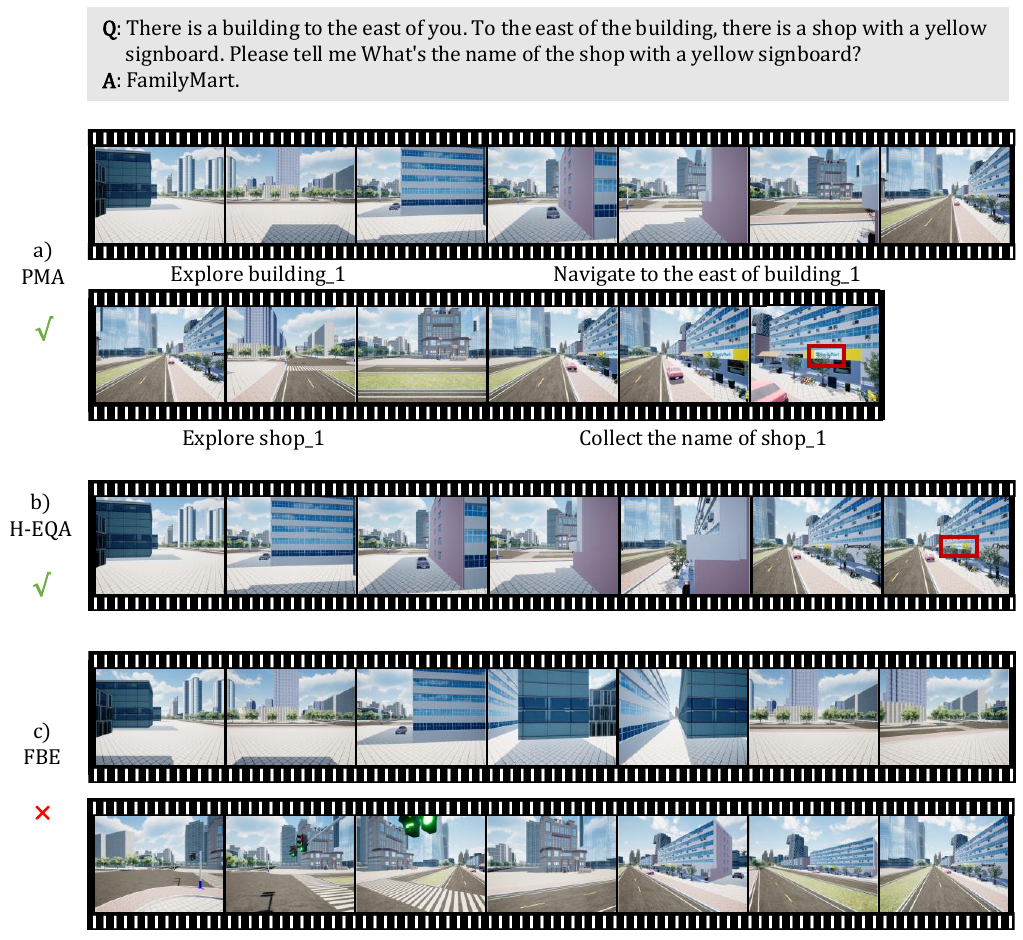}
\caption{Examples of different EQA methods.}
\label{fig:explorecase}
\end{figure*}

\paragraph{Baselines Details}

This section provides additional details for the baselines.

\begin{itemize}[leftmargin=*]

    \item \textbf{Blind Agents.} \zy{We choose four State-of-the-Art LLMs as blind agents, including GPT-4, Qwen-2.5, LLaMA-v3.1-8b, and DeepSeek-v3. They generate the answer purely based on the question, formulated as $\hat{y}=\text{LLMs}(q)$.}

    \item \textbf{Socratic Agents.} \zy{We sample efficient trajectories generated by H-EQA to simulate the observations available to Socratic Agents. Specifically, we select the last five frames from each trajectory and use GPT-4o to generate image captions $C$. Different LLMs—including GPT-4, Qwen-2.5, LLaMA-v3.1-8b, and DeepSeek-v3—are then used to produce the final answers, formulated as $\hat{y}=\text{LLMs}(q,C)$.}
    
    \item \textbf{VQA Agents.} \zy{They have direct access to images containing the answers. We use GPT-4o, Qwen-2.5, and LLaVA-v1.5-7b as VLMs to generate answers based on the images and questions, formulated as $\hat{y}=\text{VLMs}(q,p^{obs})$.}

    \item \textbf{Exploring Agents.} \zy{They are guided by different exploration strategies such as RE and FBE, and generate the answer based on the visual input $I$ at the termination position, formulated as $\hat{y}=\text{VLMs}(q,I)$. RE randomly selects an action from \{\textit{MoveForward}, \textit{TurnLeft}, \textit{TurnRight}, \textit{Stop}\} at each step. The angles for \textit{TurnLeft} and \textit{TurnRight} are set at 30°, and the distance for \textit{MoveForward} is 10 meters, consistent with the setting of the Navigator in the PMA. FBE identifies the frontiers between explored and unexplored regions, samples one as the navigation point, and employs the A* algorithm to find a path. The maximum path length is also limited as 10 meters. To avoid excessive exploration, GPT-4o is employed to decide when to stop.}

    \item \textbf{Human Agents.} \zy{At each step, H-EQA can only access the RGB image of the current pose and must choose one action from \{\textit{MoveForward}, \textit{TurnLeft}, \textit{TurnRight}, \textit{Stop}\}. The angles for \textit{TurnLeft} and \textit{TurnRight} are set at 30°. When selecting \textit{MoveForward}, the agent must also provide an integer distance within 10 meters. When choosing \textit{Stop}, the H-EQA is required to provide the answer.}
    
\end{itemize}

\paragraph{\zy{Categroy-level performance of the PMA}}
\zy{The category-level performance of the proposed PMA is shown in Figure \ref{fig:category}, and it varies across task types. PMA achieves the highest QAA on World Knowledge  tasks, likely because these tasks rely partially  on the LLM’s inherent knowledge and require  minimal visual inputs. However, it performs the worst on Object Recognition tasks due to their open-ended answers and greater reliance on visual inputs.}

\paragraph{\zy{Comparison between different EQA methods}}
\zy{We present the trajectories of PMA, H-EQA, and FBE to illustrate the different strategies adopted by them when searching for the answer to the same question, as shown in Figure \ref{fig:explorecase}. PMA finds the answer by decomposing the perception process into several sub-tasks and completing them step by step. H-EQA, with its stronger visual understanding and spatial reasoning abilities, can locate the answer in fewer steps. Moreover, H-EQA is often able to determine the answer from a greater distance, likely due to its extensive world knowledge, which allows it to fill in missing information even with incomplete observations. In contrast, FBE, lacking the ability to utilize landmarks such as building\_1 and shop\_1, can only fully explore the environment, resulting in lower perception efficiency. This highlights the differences between performing EQA tasks in urban spaces versus indoor environments.}

\begin{figure}[!htb]
\centering
    \includegraphics[width=\linewidth]{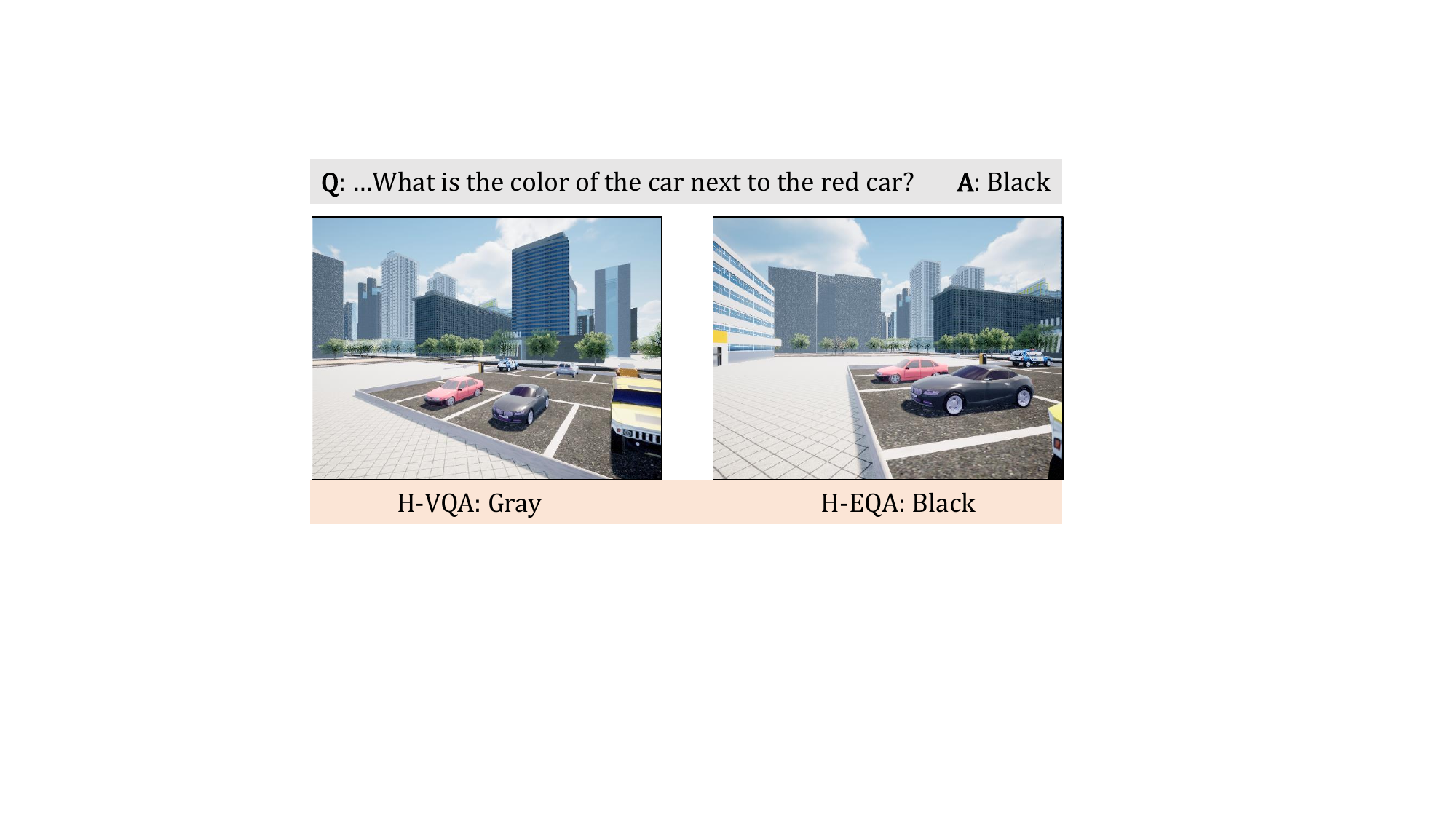}
\caption{\zy{Examples of the H-VQA and H-EQA.}}
\label{fig:case1}
\end{figure}

\paragraph{\zy{Comparison among Human Agents}}
\label{a_baseline}

\zy{In Figure \ref{fig:case1}, we provide a case to illustrate why the performance of H-EQA is superior to that of H\-VQA. The given question is "What is the color of the car next to the red car?" The ground truth answer is "Black". H\-VQA was provided with the RGB image on the left for question answering. However, in this image, due to the influence of outdoor lighting, the originally black car appears gray, thus H-VQA provided an incorrect answer. In contrast, H-EQA can actively adjust the observation pose, observing the side of the car to reduce the impact of the lighting, and thereby providing the correct answer.}

\begin{figure}[!htb]
\centering
    \includegraphics[width=\linewidth]{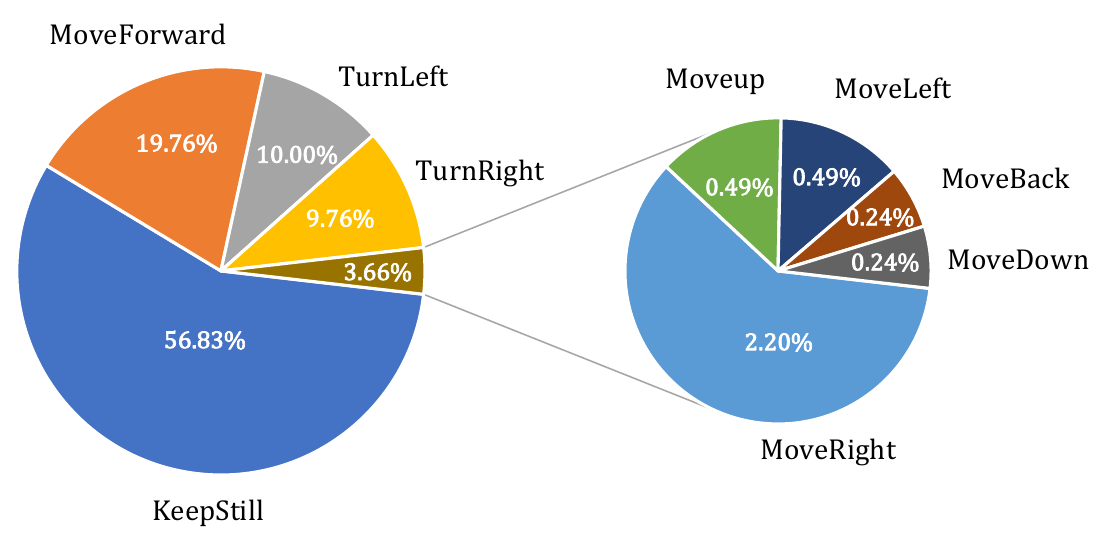}
\caption{The proportion of different actions taken by Collector.}
\label{fig:action_count}
\end{figure}

\begin{figure}[!htb]
\centering
    \includegraphics[width=\linewidth]{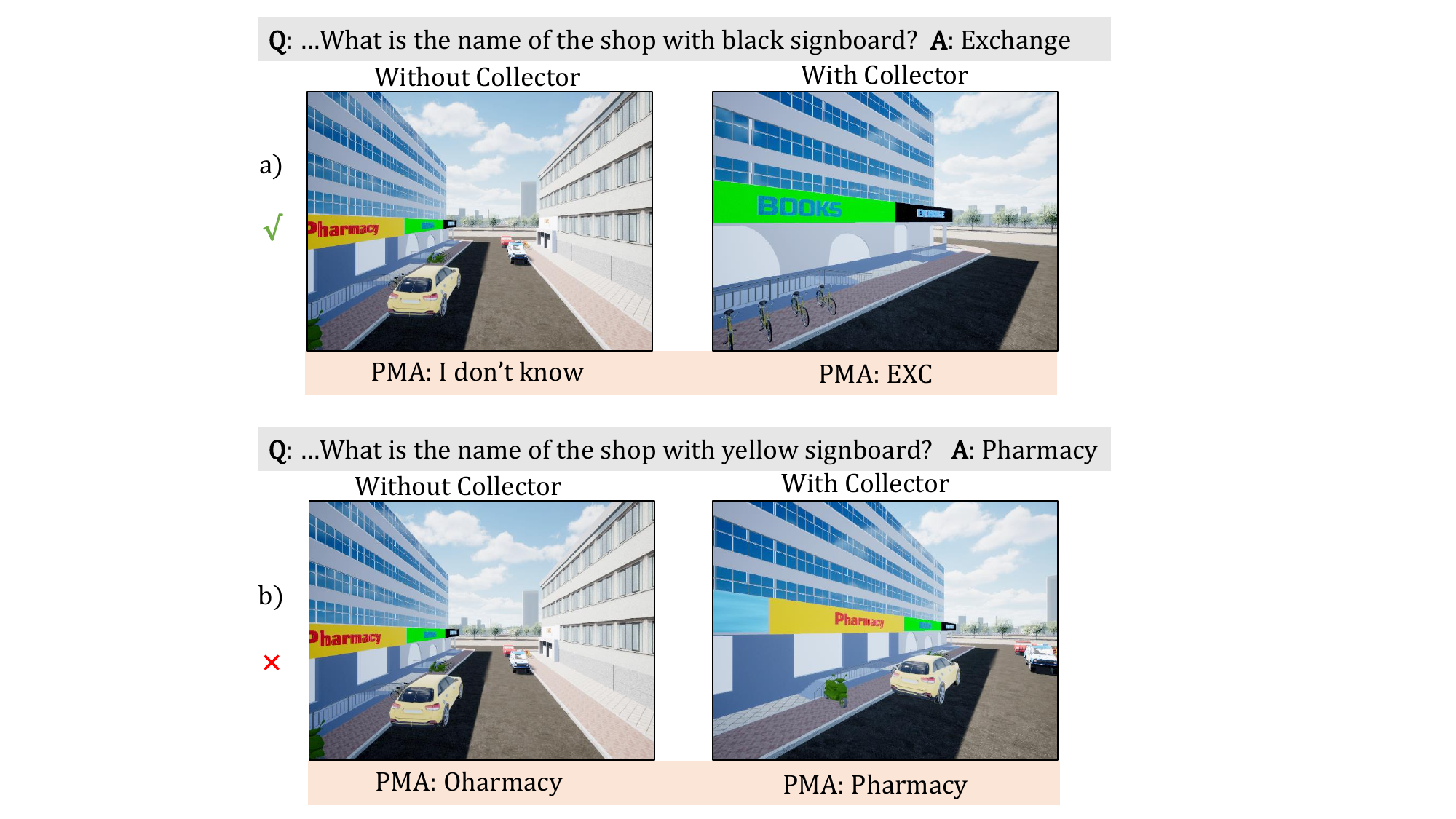}
\caption{Examples of the Collection phase.}
\label{fig:case2}
\end{figure}

\paragraph{\zy{Analysis of Collector's action}}
\label{collector_action}
\zy{The statistics of various actions taken by Collector are shown in Figure \ref{fig:action_count}. Besides, we present two cases to illustrate the effect of the collector. In the first case, as shown in Figure \ref{fig:case2} (a), since the shop with black signboard was discovered too early in the Exploration stage, the starting pose of the collector was far from the target pose. Even after moving 10 steps promptly, it still failed to recognize the text on the black signboard. In the second case, as shown in Figure \ref{fig:case2} (b), the yellow signboard that the collector needed to recognize was on the left side of the picture and seemed not to be fully displayed. At this time, the collector took the \textit{TurnLeft} action, thus observing the entire yellow signboard and easily providing the correct answer.}


\end{document}